\documentclass[9pt]{IEEEtran}
\usepackage{graphicx}
\usepackage{tikz}
\usepackage{amsmath}
\usepackage{amssymb}
\usepackage{times}
\usepackage{float}
\usepackage{todonotes}
\usepackage{pgfplots}
\usepackage{subcaption}
\usepackage{color}
\usepackage{wrapfig}

\begin{document}
\sloppy

\pagestyle{plain}

\title{ClosNets: a Priori Sparse Topologies for Faster DNN Training}

\author{Mihailo Isakov and Michel A. Kinsy\\
Adaptive and Secure Computing Systems Laboratory \\
Department of Electrical and Computer Engineering\\
Boston University, Boston, USA
\vspace{-0.4in}}

\maketitle
\begin{abstract}
    Fully-connected layers in deep neural networks (DNN) are often the throughput and power bottleneck during 
    training. This is due to their large size and low data reuse. 
    Pruning dense layers can significantly reduce the size of these networks, but this approach can only be applied
    after training.
    In this work we propose a novel fully-connected layer that reduces the memory requirements of DNNs
    without sacrificing accuracy. 
    We replace a dense matrix with products of sparse matrices whose topologies we pick in advance. This allows us to:
    (1) train significantly smaller networks without a loss in accuracy, and (2) store the network weights without
    having to store connection indices. We therefore achieve significant training speedups 
    due to the smaller network size, and a reduced amount of computation per epoch.
   We tested several sparse layer topologies and found that Clos networks perform well due to their high path
    diversity, shallowness, and high model accuracy. With the ClosNets, we are able to reduce dense layer sizes by as
    much as an order of magnitude without hurting model accuracy. 
\end{abstract}

\vspace{-0.2in}
\section{Introduction}
Training deep neural networks (DNN) is both time-consuming and power intensive. While the bulk of the computation in applications like image processing revolves around convolutional layers, this is not the case for many other applications, e.g., text-to-speech processing, machine translation and financial forecasting. Fully-connected layers, i.e., dense layers, are more common in those applications since their data exhibit more temporal correlations. Consequently, their computation requires a much larger memory footprint, more data movement, and longer processing time with significantly more power~\cite{DBLP:journals/corr/HanMD15}.
Several works have attempted to decrease the size of dense layers by pruning~\cite{DBLP:journals/corr/HanMD15} or
quantization~\cite{DBLP:journals/corr/CourbariauxB16}. However, most of these methods only speed up inference and not training. 
In this work we tackle the challenge of speeding up deep neural network training by decreasing the size of these dense layers. 
To achieve this we break up the dense matrices into products of sparse matrices. These products retain full
connectivity while requiring less parameters.  Our work is orthogonal to quantization
and can further benefit from approaches such as ZipML~\cite{DBLP:journals/corr/ZhangK0A0Z16}. 

\begin{figure}[h]
    \centering
    \includegraphics[width=0.35\textwidth]{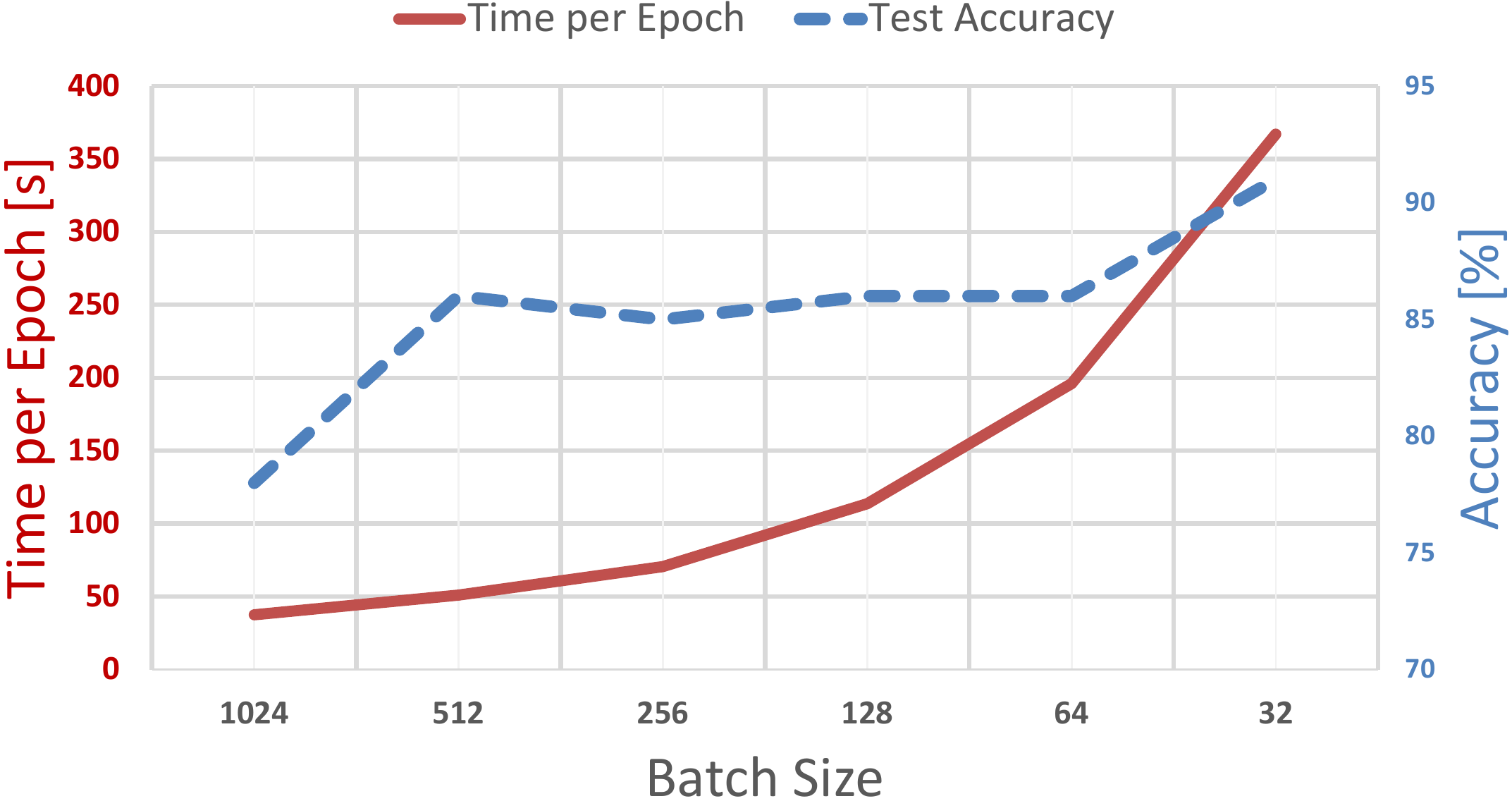}
    \caption{Time required to train a ResNet18 on a single epoch of the CIFAR-10 dataset, with varying batch sizes.}
    \label{fig:batches}
    \vspace{-0.2in}
\end{figure}

To speed up the training of dense neural networks, we investigate the bottlenecks of accelerating dense layers.
In order to determine whether networks are computationally or memory bound, we measure the time required for training
on one epoch with varying batch sizes. While the number of operations required to train on an epoch is independent of
the batch size, the number of times we have to load all weight matrices is equal to the number of batches. 
It can be noted in Figure~\ref{fig:batches} that the training time grows inversely with the batch size.
We attribute this effect to the system being computationally bound for large batch sizes and memory bound for small batch 
sizes with increased number of batches. 
To reduce the impact of this memory wall, one can either increase the bandwidth or decrease the amount of memory required
for training. Here we focus on decreasing the DNN memory requirements.
We explore both algorithmic modifications to the neural network structure and hardware customization techniques to reducing 
the size of the dense layers.

In this work we introduce the concept of predefined topology for sparse neural networks to enable faster inference and training, along with lower memory requirements and power usage. This predefined topology needs to have the following properties: (1) full connectivity, (2) shallowness, (3) pre-determined connectivity, (4) uniform and high path diversity, and (5) an efficient hardware implementation.

\vspace{-0.2in}
\section{ClosNets}
\begin{figure}[b]
\vspace{-0.1in}
  \begin{subfigure}[b]{0.23\textwidth}
    \includegraphics[width=\textwidth]{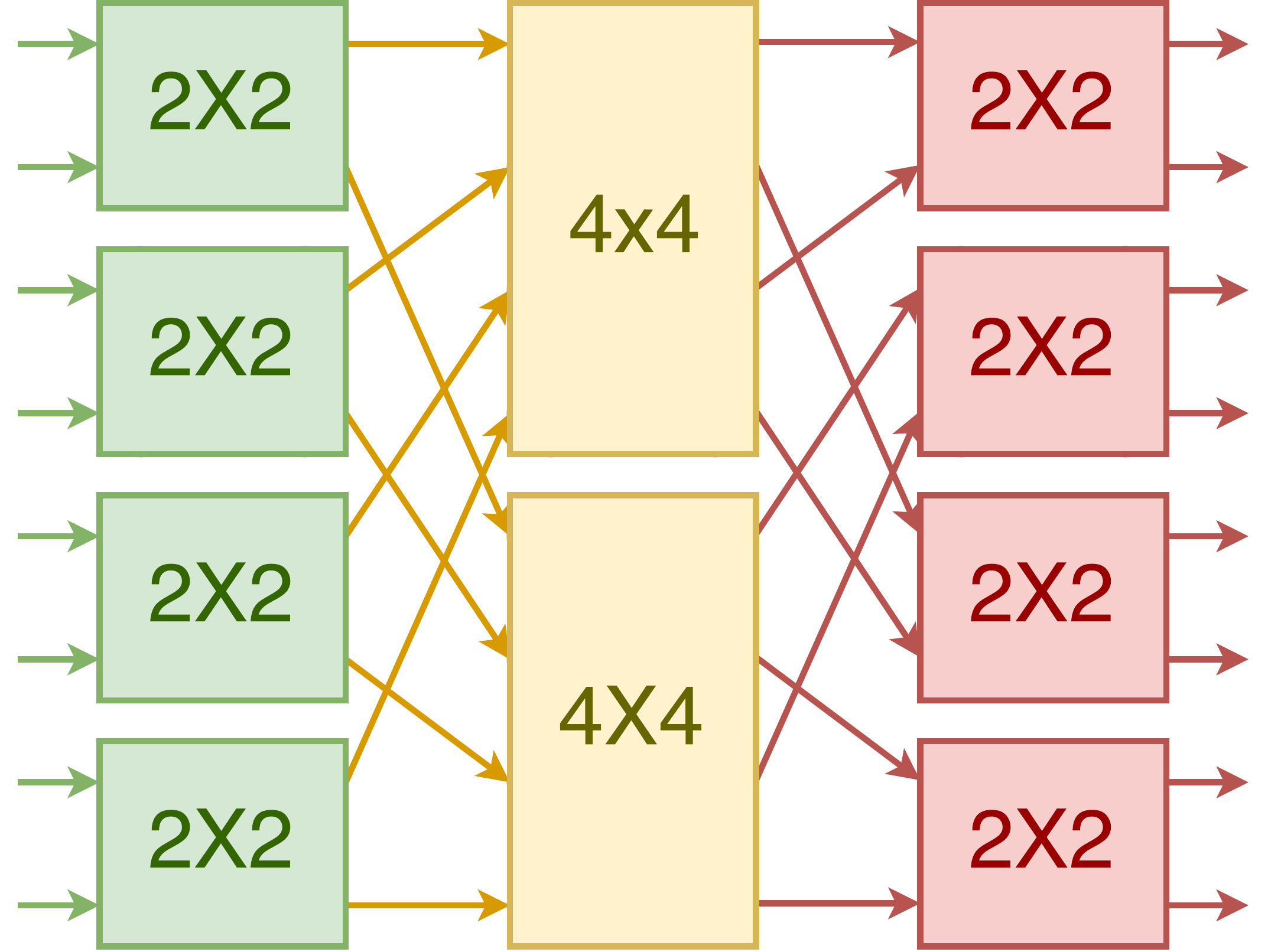}
    \caption{A 16-input, 16-output Clos network with 4 input routers, 2 middle routers, and 4 output routers.}
    \label{fig:clos_routers}
  \end{subfigure}
  \hfill
  \begin{subfigure}[b]{0.23\textwidth}
    \includegraphics[width=\textwidth]{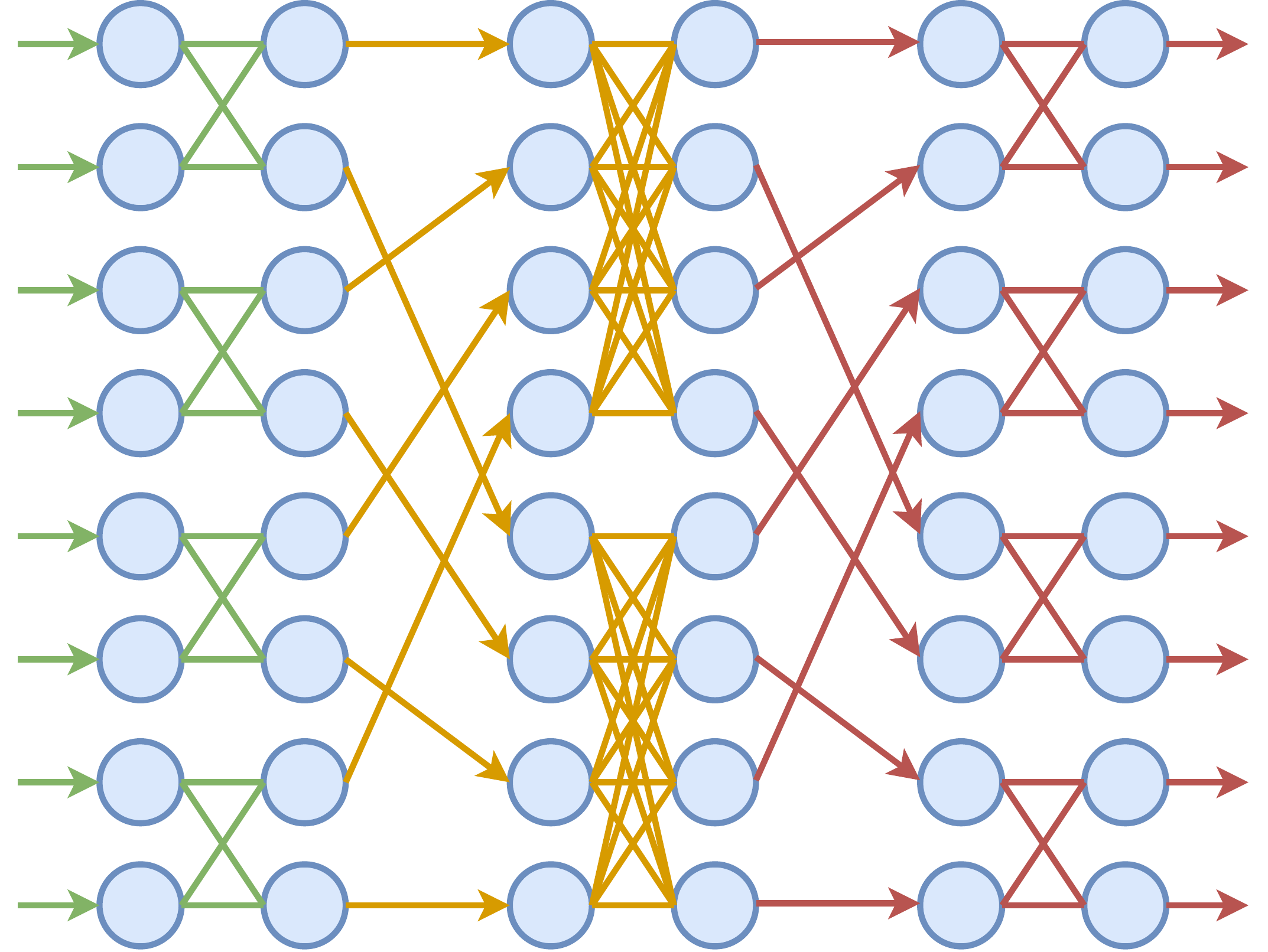}
    \caption{The same network from Figure~\ref{fig:clos_routers} mapped to the neural network domain.}
    \label{fig:clos_nn}
  \end{subfigure}
  \caption{Clos in the networking and DNN domain.}
\end{figure}
In the search for a structure that meets the above mentioned properties of a predefined topology, we examine several topologies, e.g., torus, hypercube, butterfly and mesh. Most of them do not satisfy the (2) shallowness
requirement, as they require many cascading layers before (1) full connectivity is achieved. One topology 
that grants all of these properties is \textit{Clos network}. 
A Clos network is a three-stage network in which each stage is composed of a number of crossbar
switches~\cite{Dally:2003:PPI:995703}.
While in the networking domain a Clos network is assumed to have the same number of input and output nodes, we define a
more general Clos network as a 5-tuple ($I$, $O$, $R_i$, $R_m$, $R_o$). In this characterization, $I$ is the number of
inputs, $O$ is the number of outputs, $R_i$ is the number of input routers, $R_m$ is the number of middle routers, and
$R_o$ is the number of output routers.
In Figure~\ref{fig:clos_routers} we can see an example (16, 16, 4, 2, 4) Clos network.

The intuition behind Clos networks is that since each middle router acts as a crossbar between the input and output
routers, there are as many paths between an input-output pair $(i, o)$ as there are middle routers. This gives the network
designer a simple way of preventing network contention by increasing the path diversity.
To map Clos networks to the DNN domain, we replace all the input and output nodes with
neurons, and each router becomes a fully-connected network of the same size with its own hidden neurons. It is important
to note that the connections between the routers are a simple scatter operation. These connections do not have weights
(i.e. they do not amplify or inhibit their signals), but purely permute the positions of the activations.
In Figure~\ref{fig:clos_nn} the network from Figure~\ref{fig:clos_routers} is mapped to the neural network domain. 

The Clos neural networks have clear benefits over the other explored topologies. They are fully-connected and shallow. They 
have a parametrizable but uniform path diversity. Furthermore, they offer a simple hardware implementation.
An added benefit of Clos networks is that we are not restricted to having either (1) the identical number of inputs and
outputs, or (2) number of inputs/outputs being a power of two. 
For a given Clos network ($I$, $O$, $R_i$, $R_m$, $R_o$), we can calculate the number of parameters $P$ in the network,
and path diversity $D$ as:
\begin{equation}\label{eq:parameters}
\begin{split}
    P = R_m(I + O + R_i R_o),\quad D=R_m
\end{split}
\end{equation}

\begin{figure}[h]
    \centering
    \includegraphics[width=0.4\textwidth]{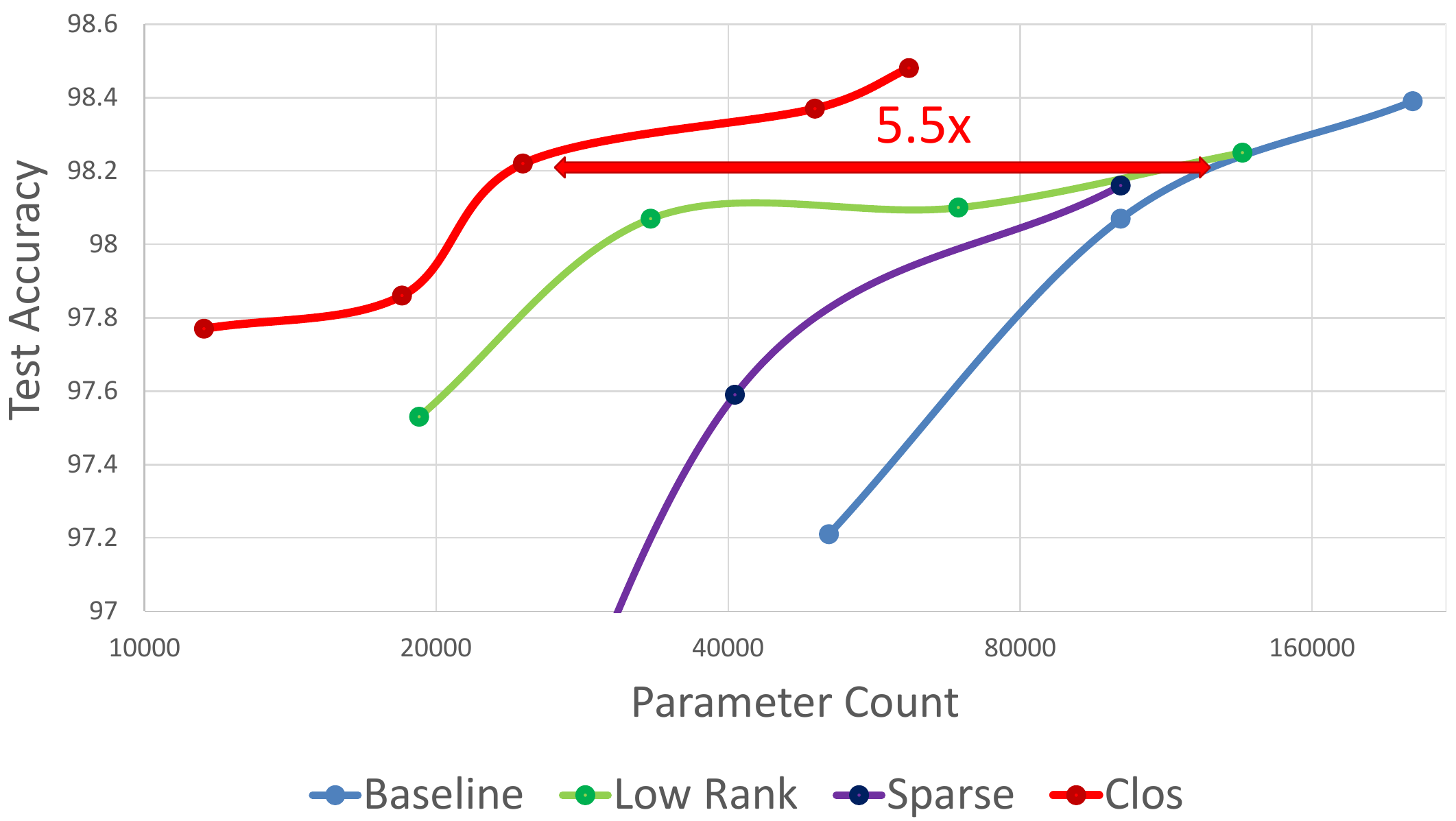}
    \caption{Accuracy vs. parameters tradeoff for different network types.}
    \label{fig:param_vs_acc}
\end{figure}

We compare the model accuracies of different configurations of the baseline (dense) models, low-rank models, a priori
pruned models and Clos models, with respect to the number of parameters of each model. 
Figure~\ref{fig:param_vs_acc} shows the test accuracy of different networks trained on MNIST for 50
epochs with respect to their parameter count. As we reduce the parameter count, the networks degrade in accuracy. 
The performance degradation is more graceful for some networks compared to others. Notice that 
the Clos networks have comparable accuracy with the baseline networks while having $5.5\times$ less parameters.

\vspace{-0.2in}
\section{Architecture}
\begin{figure}[h]
    \centering
    \includegraphics[width=0.44\textwidth]{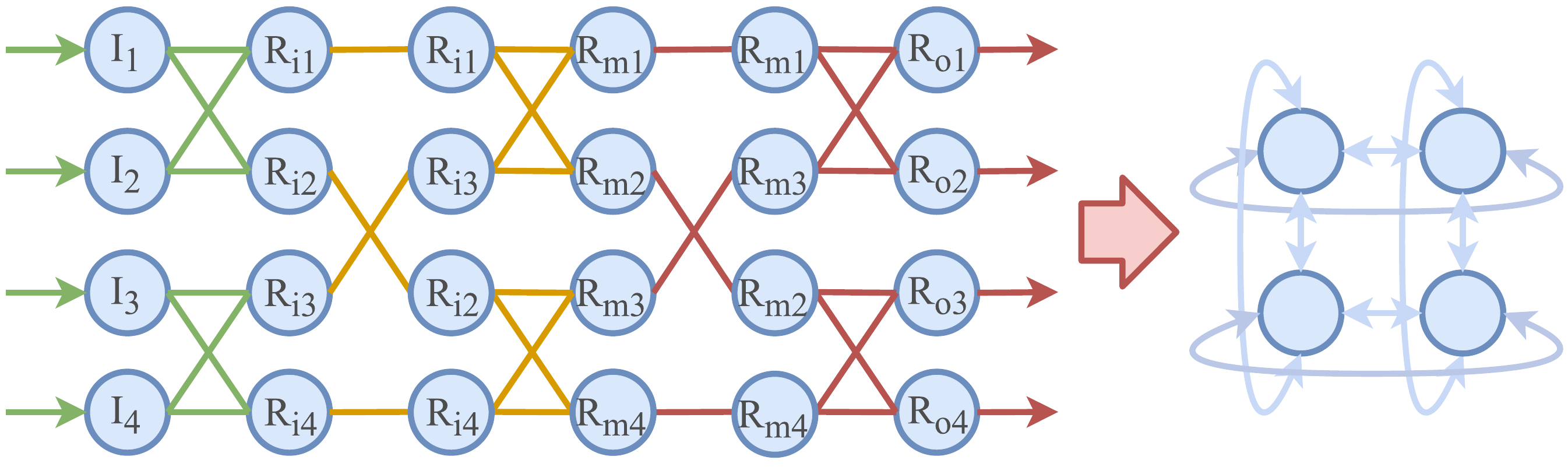}
    \caption{(Left) A (4, 4, 2, 2, 2) Clos network with inputs $I$, intermediate results $R_i$ and $R_m$, and outputs $R_o$.
    (Right) A $2\times 2$ torus we want to map the Clos network to.}
    \label{fig:torus_top}
\end{figure}

\begin{figure}[h]
    \centering
    \includegraphics[width=0.44\textwidth]{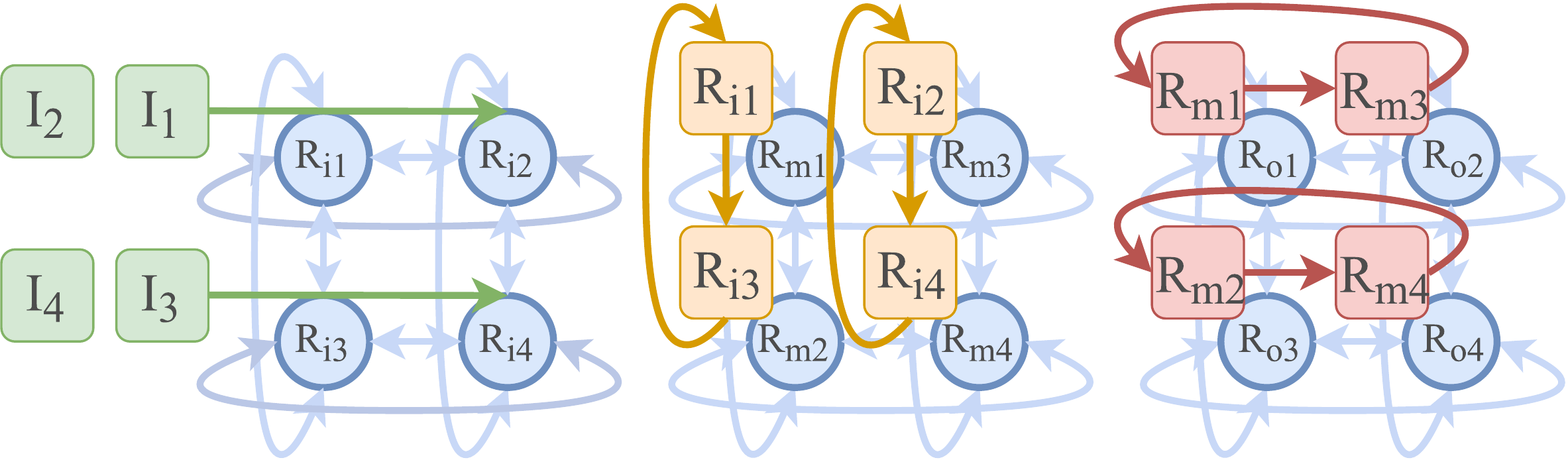}
    \caption{The three rings router outputs from the Clos network calculated with a ring-AllReduce operation.}
    \label{fig:torus_bottom}
    \vspace{-0.2in}
\end{figure}

\begin{figure}[h]
    \vspace{-0.1in}
    \centering
    \includegraphics[height=0.15\textwidth]{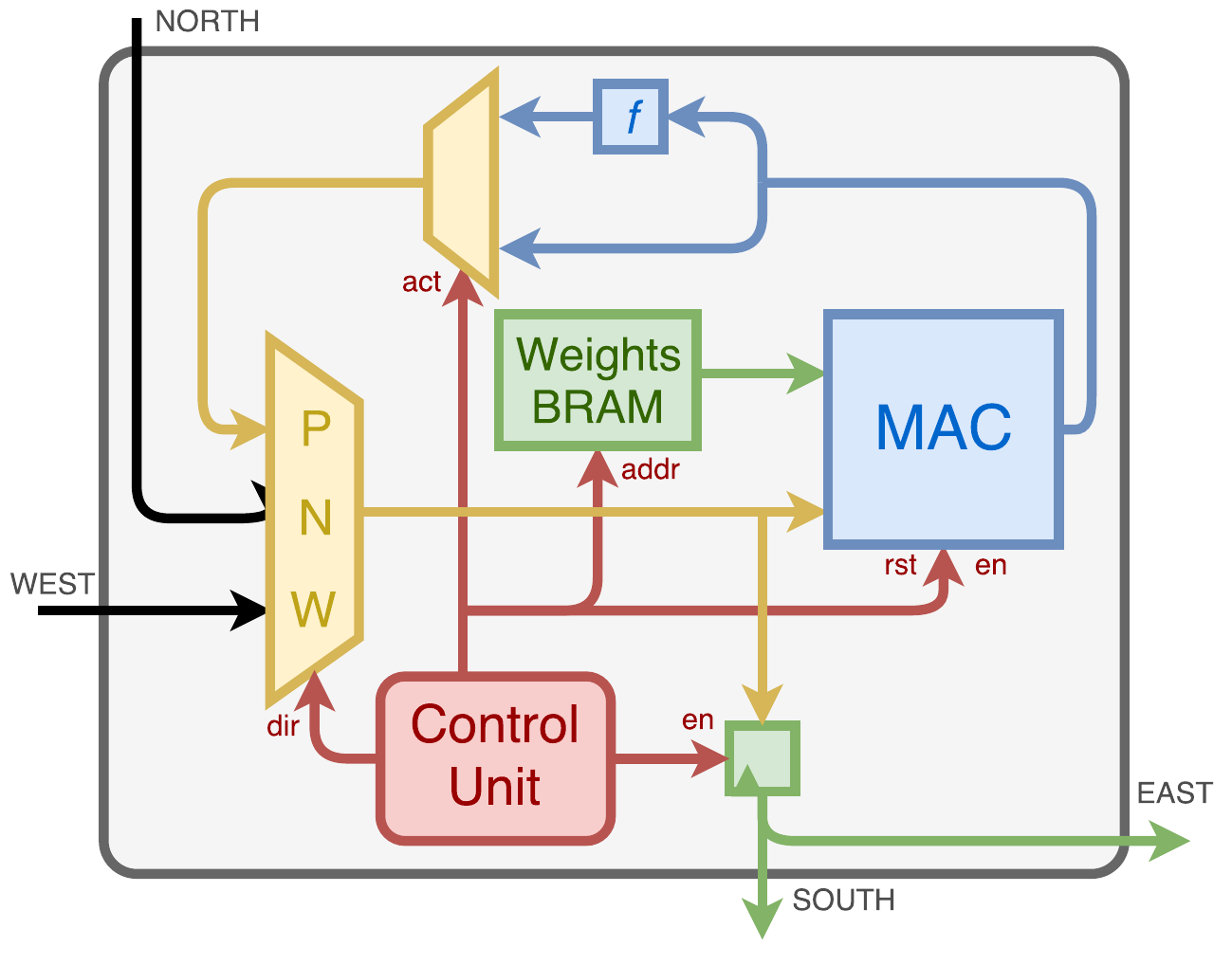}
    \vspace{-0.1in}
    \caption{Inference router, accepting inputs from north and west, and forward them east and south. Once the router
    computes the neuron activation, it sends it south or east.}
    \label{fig:inference_router}
    \vspace{-0.1in}
\end{figure}

\begin{figure}[h]
   \vspace{-0.1in}
    \centering
    \includegraphics[height=0.15\textwidth]{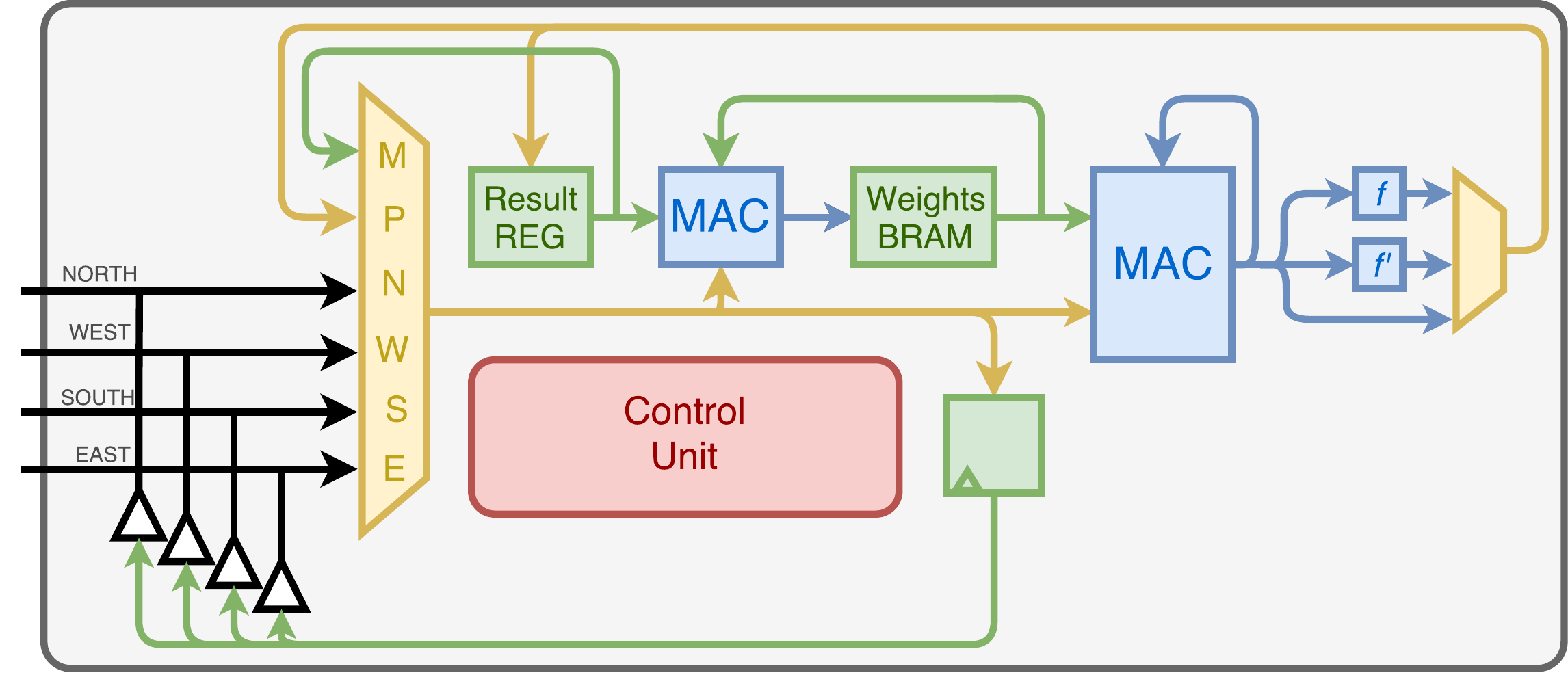}
    \caption{A training router, with bidirectional links to all four directions. The router processes both forward and backward passes. }
    \label{fig:backprop_router}
    \vspace{-0.1in}
\end{figure}

We propose a simple hardware implementation for our Clos networks. For brevity, we will restrict ourselves to networks
with the same number of inputs and outputs, and optimize for throughput, not area or latency.
By observing Figure~\ref{fig:clos_nn}, we can see that the network consists of several independent fully-connected
layers, connected by a scatter operation. We can implement the fully-connected layers with a number of processing elements
connected in a ring topology. Each processing element is calculating the value of one of the output neurons of the dense
layer, with the input neuron values circling the ring. Once each ring has calculated the outputs of each neuron in a
single column, the torus reconfigures to a set of perpendicular rings, one for each column in this case.
Each torus node now calculates a new neuron output from the next column from Figure~\ref{fig:clos_nn}, while the
previous results are treated as inputs and circle the rings.
We illustrate this on a small $4\times 4$ Clos network as seen in Figure~\ref{fig:torus_top}.
We map this network to the $2\times2$ torus from Figure~\ref{fig:torus_top} (right). 
This leads us to a torus implementation as seen in Figure~\ref{fig:torus_bottom}. The three router layers (input, middle,
output) then map to three ring-AllReduce operations, a horizontal one over each row individually (green), a vertical one over
each column (yellow), and again a horizontal one over each row (red). Notice that the positions of the neurons
$R_i, R_m$ and $R_o$ change when mapped to torus nodes (positions 2 and 3 swap places). This is done to minimize data
movement and allow single hop movement of neuron outputs.
From figure~\ref{fig:torus_bottom}, we see that neuron outputs are transmitted only in the east and south directions.
This is true only for inference. During training, gradients backtrack their path through the network, flowing north and
west. Therefore, we implement two router designs, one for inference -- Figure~\ref{fig:inference_router} and the other for training -- Figure~\ref{fig:backprop_router} to support the communication patterns described above. 

\vspace{-0.1in}
\section{Conclusion}
In this work we introduce a novel approach for reducing the size of dense neural network layers. 
We present ClosNets - fully-connected cascades of sparse layers with the Clos topology.
We show that ClosNets have comparable accuracy and $5.5\times$ smaller size over conventional fully-connected layers, and propose a simple
torus-based implementation of the network.
\vspace{-0.1in}

\bibliographystyle{IEEEtran}
\bibliography{paper}


\end{document}